# Analysis of Convolutional Neural Network-based Image Classifications: A Multi-Featured Application for Rice Leaf Disease Prediction and Recommendations for Farmers

Biplov Paneru[1], Bishwash Paneru[2], Krishna Bikram Shah[3]
[1]Department of Electronics & Communication Engineering, Nepal Engineering College Pokhara University, Nepal
[2]Department of Applied Sciences and chemical Engineering, Tribhuvan University, Institute of Engineering, Nepal
[3]Department of Computer Science and Engineering, Pokhara University, Nepal
Corresponding author: First A. Author (e-mail: biplovp019402@nec.edu.np)

**ABSTRACT** This study presents a novel method for improving rice disease classification using 8 different convolutional neural network (CNN) algorithms, which will further the field of precision agriculture. A thorough investigation of deep learning methods is carried out using the UCI dataset in order to create a reliable and effective model that can correctly identify a range of rice diseases. The suggested transfer learning models perform better at identifying subtle features and complex patterns in the dataset, which results in extremely accurate disease classification. Moreover, the study goes beyond the creation of models by incorporating an intuitive Tkinter-based application that offers farmers a feature-rich interface. With the help of this cutting-edge application, farmers will be able to make timely and well-informed decisions by enabling real-time disease prediction and providing personalized recommendations. Together with the user-friendly Tkinter interface, the smooth integration of cutting-edge CNN transfer learning algorithms-based technology that include ResNet-50, InceptionV3, VGG16, and MobileNetv2 with the UCI dataset represents a major advancement toward modernizing agricultural practices and guaranteeing sustainable crop management. Remarkable outcomes include 75% accuracy for ResNet-50, 90% accuracy for DenseNet121, 84% accuracy for VGG16, 95.83% accuracy for MobileNetV2, 91.61% accuracy for DenseNet169, and 86% accuracy for InceptionV3. These results give a concise summary of the models' capabilities, assisting researchers in choosing appropriate strategies for precise and successful rice crop disease identification. A severe overfitting has been seen on VGG19 with 70% accuracy and Nasnet with 80.02% accuracy. On Renset101, only an accuracy of 54% could be achieved, along with only 33% on efficientNetB0. A MobileNetV2-trained model was successfully deployed on a TKinter GUI application to make predictions using image or real-time video capture.

## I. INTRODUCTION

Rice is by far the most significant food crop for people in low- and lower-middle-income countries out of the three main crops: wheat, maize, and rice. Rice is a basic and typically indispensable food staple in many Asian nations, particularly for the impoverished [1]. Rice makes up approximately half of the food expenses and, on average, a fifth of the total household expenditures for the extreme poor of Asia, who survive on less than $1.25 per day. By purchasing power parity, this group alone spends $62 billion a year on rice. A lot of the world's impoverished people depend on rice for their food security. Because it grows well in a variety of environments and is high in proteins and carbohydrates, rice is the most widely grown crop in India. As a result, over 1.2 billion people in South Asia and an estimated 3.5 billion people globally use rice as their main meal [2]. Rice is the main source of energy for over half of the world's population [3]. Numerous farmers worldwide are impacted by illnesses associated with rice leaves, which can seriously jeopardize the sustainable production of rice [20].
 To reduce yield losses, rice infections must be identified and treated as soon as possible. Plant disease identification has shown considerable promise for convolutional neural networks (CNNs). However, CNN training requires enormous datasets of annotated images, which can be costly and time-consuming [4]. Today's agricultural industry is poised for a technological revolution that will necessitate a paradigm change toward precision farming practices. The introduction of AI and machine learning into agriculture has the potential to bring about previously unattainable breakthroughs in yield optimization, disease detection, and crop management. Our work aims to make a substantial contribution to this revolutionary wave by concentrating on the use of transfer learning algorithms in the field of rice disease classification.

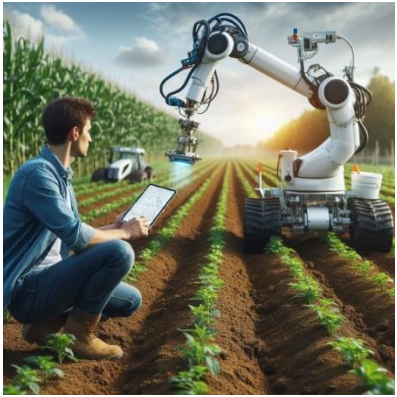

**Figure 1.** AI and technologies in agriculture

The integration of AI and machine learning emerges as a ray of hope as traditional farming methods struggle to meet the demands of a sustainable agricultural system as shown in figure 1. Our research seeks to transform the rice cultivation landscape through the utilization of sophisticated algorithms and data-driven insights. The implementation of CNNs, especially with the use of the UCI dataset, represents a significant advancement in tackling the complex problems that farmers encounter in crop management and disease detection. We're more than just interested in creating the transfer learning-based models but cutting edge application is developed by integrating the model. We understand that the agricultural community needs solutions that are realistic, approachable, and practical. In order to achieve this, we have created a Tkinter-based application that works in unison with our CNN model. This dynamic interface offers farmers personalized recommendations in addition to its unmatched accuracy in predicting rice diseases. Our research presents a comprehensive and holistic solution that has the potential to revolutionize agriculture and pave the way for a more sustainable and efficient future. This is achieved through the convergence of AI-driven disease classification and a user-centric Tkinter application.

## II. LITERATURE REVIEW

Given the similarities in appearance between some types of rice disease, an overall accuracy of 91% was obtained when six types of rice diseases were diagnosed using the Ensemble Model. This is regarded as reasonably good. Using the smartphone app, the client could easily and effectively diagnose rice leaf blast, false smut, neck blast, sheath blight, bacterial stripe disease, and brown spot in the field. The Ensemble Model was accessible on the web server via a network [1].

Paddy crops must be protected, and early disease detection is essential. In the past, diagnosing a disease required observation or laboratory examination. Visual observations require expertise and can differ from person to person, making them prone to error. Laboratory testing also takes longer and may not yield results quickly. Machine learning techniques based on image processing were employed to identify and categorize the diseases in order to overcome this problem. Authors primarily addressed diseases of rice (Oryza sativa) [2].

In order to obtain the reduced data with significant features that are used as the input to the classification model, the feature extraction process is first applied to the data in this work. Following this, feature selection is also applied. Color, shape, position, and texture features are extracted from photos of infected rice plants for the rice disease datasets, and a rough set theory-based feature selection technique is applied to the feature selection task. In order to develop an efficient disease prediction model for the classification task, ensemble classification methods have been implemented within a map reduce framework. The effectiveness of the suggested model is demonstrated by the findings on the gathered disease data [3].

Using a dataset of 8883 and 1200 photos of diseased and healthy rice leaves, respectively, the suggested method was tested and found to achieve an accuracy of 94% using the 10-fold cross-validation process, which was significantly higher than other approaches. These simulation results demonstrate the viability and effectiveness of rice disease detection and provide an affordable, easily accessible means of identifying rice diseases early on. This is especially helpful in developing nations with limited resources and can make a substantial contribution to the production of sustainable food [4].

The three most common diseases that affect rice leaves are brown spot, leaf blast, and hispa. In order to address this problem, we have investigated a number of deep learning and machine learning techniques for identifying the diseases on their leaves. We have measured the effectiveness of these techniques by calculating their accuracy, recall, and precision. By identifying diseases in rice leaves, this study assists farmers in obtaining a healthy crop yield. When compared to machine learning techniques, the deep learning models exhibit superior performance. After examining every deep learning model, we discovered that the 5-layer convolution model performed the best, with an accuracy of 78.2%, while other models, like VGG16, performed worse, with an accuracy of 58.4% [5].

An automated diagnosis technique was created for this study and put into a smartphone app. Based on a sizable dataset of 33,026 photos of six different forms of rice diseases—leaf blast, false smut, neck blast, sheath blight, bacterial stripe disease, and brown spot—the technique was created using deep learning. According to the results, the top three submodels in terms of a number of characteristics, including learning rate, precision, recall, and accuracy in identifying diseases were DenseNet-121, SE-ResNet-50, and ResNeSt-50. An overall accuracy of 91% was obtained when six different types of rice illnesses were diagnosed using the Ensemble Model. The client was able to use the Ensemble Model on the web server over a network thanks to the smartphone app, which made it easy and effective for field diagnosing illnesses [6].

A device that can automatically detect issues with rice plants is desperately needed by the agriculture industry. We introduce a new convolutional neural network (CNN) model that may be used to classify common illnesses of rice leaves. Our program can detect rice leaf diseases from a range of image backgrounds and capture scenarios. Our objective is to classify disease images in rice leaves with complex backgrounds and different lighting conditions. Our accuracy using the CNN-based model is 95%. The results for rice disease identification show how successful the recommended strategy is. Index terms include machine learning, CNN algorithm, rice leaf, and disease detection [7].

The automatic identification of pathogenic organisms in the leaves of rice plants is made possible by the significant advancements in agricultural technology. One of the deep learning algorithms that has been successfully used to solve computer vision issues such as picture classification, object segmentation, image analysis, etc. is the convolutional neural network algorithm (CNN). InceptionResNetV2, a kind of CNN model, is used in this work along with a transfer learning strategy to identify illnesses in photos of rice leaves. The suggested model's parameters were fine-tuned for the classification challenge, yielding a respectable 95.67% accuracy [8].

Initially, we extract features from photos of rice leaf disease in this study. Then, we used a variety of machine learning approaches to categorize the photos, and we discovered that the Quadratic SVM classifier produced an accuracy of 81.8%. Different forms of rice diseases were also distinguished from one another using shape criteria like area, roundness, and area to lesion ratio, among others. The outcomes were satisfactory and fulfilled the necessary requirements [9].

In this publication, a CNN-VGG19 model—a convolutional neural network and visual geometry group—with a transfer learning-based approach for the accurate diagnosis and categorization of rice leaf diseases was presented. This plan uses a transfer learning method that can determine the brown spot class and is based on the VGG19. In terms of the deployment of the rice leaf disease dataset, the accuracy is 93.0%. The remaining metrics are the F1-score, sensitivity, specificity, and precision, which have corresponding values of 89.9%, 94.7%, 92.4%, and 90.5%. When compared to the baseline models that were already in place, the created technique produced better outcomes [10]. In order to obtain a healthy crop output, this study assists farmers by identifying illnesses in rice leaves. In terms of performance, the deep learning models outperform the machine learning techniques. Upon doing an analysis of every deep learning model, we discovered that the 5-layer convolution model yielded the highest accuracy of 78.2%, whilst other models, including VGG16, had lesser accuracy of 58.4% [11].

In order to identify rice leaf disease with high accuracy, a two-stage method based on AI (artificial intelligence) algorithms is suggested. Images of rice leaf diseases in the field are automatically gathered in the first stage. These image data are then divided into sets of brown spots, leaf folders, and blast leaves, in that order. The trained model is used on IoT devices to detect and classify rice leaf illnesses in the second stage, which follows the training of the YOLOv8 model using our suggested image dataset. We conduct a comparative research between our suggested method and the methods employing YOLOv7 and YOLOv5 in order to evaluate the performance of the proposed approach. The experimental results demonstrate that the accuracy of our proposed model in this research has reached up to 89.9% on the dataset of 3175 images with 2608 images for training, 326 images for validation, and 241 images for testing. It demonstrates that our proposed approach achieves a higher accuracy rate than existing approaches [12].

The body of research on the diagnosis of rice disease using deep learning and machine learning techniques has produced some impressive results in terms of accessibility and accuracy. With the impressive overall accuracy of 91% displayed by the Ensemble Model, six different rice diseases could be diagnosed with the use of a smartphone app. But there is a gap in the literature regarding the precise ensemble techniques used, which our work can fill by examining and contrasting different ensemble approaches for improved disease diagnosis. Furthermore, the reviewed studies' emphasis on feature extraction and selection strategies highlights how crucial it is to optimize these procedures for increased model efficiency.

By providing a comparative analysis of various extraction and selection techniques, our proposed research could go deeper into identifying the most informative features of image 3 type of classes, which are: Bacterial leaf Blight, Brown Spot, and Leaf Smut that are obtained from UCI machine learning directory for rice disease classification. Moreover, there exists a deficiency in the existing literature concerning the scalability of the suggested models, specifically in environments with restricted resources. To close this gap, our work concentrates on creating transfer learning models that are effective and easily obtainable, making a significant contribution to the production of food in a sustainable manner, particularly in developing countries with limited resources.

### III. METHODOLOGY

The proposed work consists of training a deep learning model with a UCI dataset for creating an accurate model for making predictions on 3 classes of data, which are 'bacteria': 0, 'brown': 1, and 'smut': 2, that are used to make the prediction on those images, along with which real-time video capturing of the image helps to make predictions on the image. The model can make predictions on the model as shown in figure 2. and predict the disease the rice leaf has been reflecting and providing suggestions to farmers regarding the implementing methods to treat the disease.

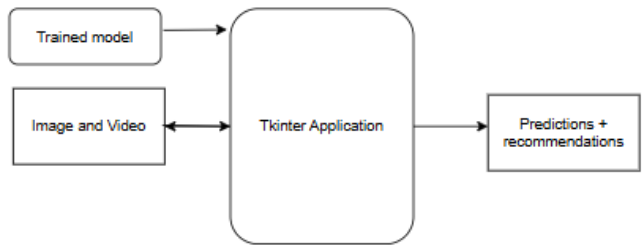

Figure 2. Proposed architecture

**ResNet-50**

Microsoft Research announced ResNet-50, also known as Residual Network with 50 layers, a deep convolutional neural network. It is well known for using residual connections, which are layer-skipping shortcuts. With the help of this architecture, the vanishing gradient issue is lessened, enabling a much deeper network without sacrificing efficiency. Because they make it easier for gradients to move across the network, residual connections aid in the training of deeper networks. ResNet-50 performs well in many image identification tasks, but because of its propensity for overfitting, it can occasionally do less well in validation accuracy, especially in smaller or less diversified datasets.

**DenseNet-121**

DenseNet-121 is a DenseNet variation with dense connections between layers that was first presented by Cornell University. In a DenseNet, every layer passes its own feature maps to every other layer after receiving inputs from all layers that came before it. Because of its dense interconnectedness, features can be reused and fewer parameters are required, which frequently leads to increased efficiency and accuracy. DenseNet-121 performs exceptionally well in training and validation accuracy, suggesting robust feature learning and good generalization. It maintains great validation accuracy and nearly faultless training performance thanks to its thick connections.

**VGG16**

The Visual Geometry Group at the University of Oxford created VGG16, which is renowned for its consistent architecture and simplicity. There are three fully connected layers and thirteen convolutional layers among its sixteen weighted layers. The model is simple to use but highly effective for picture classification tasks because it makes use of deep networks and relatively small convolutional filters (3x3). Due to its tendency to overfit, VGG16 can have issues with validation accuracy even with reasonably high training accuracy, particularly on datasets with low size or diversity.

**MobileNetV2**

Google created MobileNetV2, a platform for effective model deployment on mobile and edge devices. In order to save computation time and number of parameters without sacrificing accuracy, depthwise separable convolutions are introduced. To improve efficiency and performance even more, MobileNetV2 additionally uses inverted residual structures and linear bottlenecks. The model is durable and effective at capturing characteristics while being computationally efficient, which makes it well-suited for real-time applications. This is demonstrated by the high training and validation accuracy of the model.

**Inception V3.**

Google's Inception V3 is a deep convolutional network that applies several filters in parallel at various scales by using Inception modules. This architecture enhances the model's ability to learn by enabling it to capture a variety of features. To increase performance and efficiency, Inception V3 includes a number of improvements, such as factorized convolutions and auxiliary classifiers. It may not always obtain the highest training accuracy in comparison to other models, but it shows strong validation accuracy, suggesting that it can generalize well to new data.

**EfficicentNet B0**
A convolutional neural network architecture called EfficientNetB0 is made to operate effectively on a variety of computing platforms. It was created by Google and uses a technique called compound scaling to balance the network's depth, width, and resolution. The depthwise separable convolutions used in the mobile Inverted Residual Bottleneck (MBConv) blocks, which lower computing costs, form the foundation of the architecture. EfficientNetB0 is a good fit for server and mobile applications because it maintains great accuracy at a reduced computational cost than other networks.

**ResNet-101**

Building upon the ResNet (Residual Network) architecture, ResNet-101 is a deep convolutional neural network with 101 layers. In order to mitigate the vanishing gradient issue, it adds residual blocks with skip connections that facilitate gradient flow through the network during training. The network can effectively train very deep models thanks to these skip links. Because ResNet-101 can learn complicated features without overfitting, it is widely utilized in computer vision applications and is well-known for its strong performance in image classification tasks.

**VGG19**

The deep convolutional neural network VGG19 is well-known for being easy to use and efficient when it comes to image recognition tasks. It was created by Oxford University's Visual Geometry Group (VGG) and has 19 layers total—16 convolutional layers and 3 fully connected layers. Small 3x3 convolutional filters are used throughout the architecture to help capture minute features in photos. VGG19's simple design and strong performance on massive datasets like ImageNet have made it a popular choice for benchmarking and transfer learning.

**NasNet**

A deep learning model called NASNet (Neural Architecture Search Network) was created utilizing neural architecture search methods. NASNet, a Google creation, uses an automated search procedure to find the best network designs for particular purposes. Its modular architecture allows for the customization of construction blocks to fit a range of sizes and levels of complexity. NASNet has proven to be able to find efficient network configurations using algorithmic search, resulting in state-of-the-art performance in applications like as object detection and image categorization.

**DenseNet169**
A DenseNet architecture variation recognized for its dense connectivity structure is DenseNet169. Its 169 layers and feed-forward connectivity between each layer and every other layer mitigate the vanishing gradient issue and encourage feature reuse. DenseNet169 uses dense blocks, which enable each layer to learn rich representations with less parameters by receiving additional inputs from all layers that come before it. This design retains computing efficiency while achieving great accuracy on picture classification applications.

***A. DATASET COLLECTION AND PREPROCESSING***
There are 120 JPG pictures of diseased rice leaves in this dataset. Depending on the type of disease, the images are divided into three classes. Every class contains forty images as shown in figure 3.

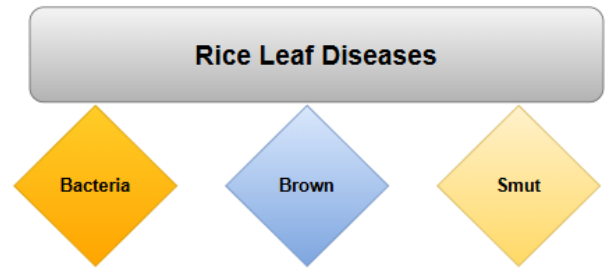

**Figure 3.** Dataset classification and breakdown

As shown in table 1. to guarantee optimal performance, the image data goes through a number of significant modifications during the preparation stage of training the MobileNetV2 model. First, pixel values are scaled to the range [0, 1] by dividing by 255 in order to normalize them. By standardizing the input data, this phase improves the effectiveness of the training process. Further augmentation methods, including as horizontal flipping, zooming, and shearing, are used on the training set to fictitiously grow the dataset and improve the model's capacity to generalize to previously unobserved images.

**Table 1.** Preprocessing steps involved

| Preprocessing Step | Description | Configuration/Values |
|---|---|---|
| **Rescaling** | Normalizes pixel values by scaling to the range [0, 1] | rescale=1./255 |
| **Data Augmentation** | Applies transformations to increase the diversity of training data | horizontal_flip=True, zoom_range=0.2, shear_range=0.2 |
| **Training Data Split** | Splits data into training and validation subsets | validation_split=0.25 |
| **Training Data Generator** | Creates batches for training with specified transformations | target_size=(img_height, img_width), batch_size=batch_size, class_mode='categorical', subset='training' |
| **Validation Data Generator** | Creates batches for validation with specified transformations | target_size=(img_height, img_width), batch_size=batch_size, class_mode='categorical', subset='validation' |
| **Test Data Preparation** | Prepares a separate dataset for evaluating the | rescale=1./255, shuffle=False |

| | model's performance | |

To assess the performance of the model during training, 25% of the dataset is put aside for validation. This divides the dataset into subsets for training and validation. These subsets are loaded and processed using different data generators: the validation data generator only rescales the input, while the training data generator applies the previously specified augmentations. Finally, to guarantee a fair assessment of the model's performance, a test set is created with the same rescaling but no augmentation. The model is trained and verified on a variety of well-prepared data thanks to this thorough preprocessing procedure, which improves accuracy and generalization.

## B. DEEP LEARNING APPROACH

Utilizing the Mobile Net V2, vGG16, RestNet50, and finally InceptionV3 models, the deep learning algorithm in the script is based on convolutional neural networks (CNNs). Finally, the InceptionV3 model showed the best results in the classification on the training and validation sets.

Setting parameters like batch size and image size, as well as dividing the dataset into training and validation sets, are all part of the training process for an ImageDataGenerator. The layers of the pre-trained InceptionV3 model on ImageNet were frozen, and a custom top classification layer was added. Categorical cross-entropy loss and the Adam optimizer are used to compile the model. The model learns from the augmented training data and validates on the separate validation set during the training loop's predetermined number of epochs. Following training, the accuracy results are printed, and the model is assessed on the training and validation sets. These measures shed light on how well the model applies to fresh, untested data.

## C. APPLICATION DEVELOPMENT

A feature-rich Tkinter GUI application is used in the study to improve the deep learning model's applicability and user-friendliness in the classification of rice diseases. This Tkinter-based interface provides a variety of functionalities to meet the various needs of farmers and agricultural practitioners, acting as a dynamic decision support system.
With its user-centric design, the GUI application facilitates seamless interaction between users and the model by accommodating multiple input modalities. A noteworthy feature allows users to upload images of rice leaves for the purpose of classifying diseases. Even those with little experience with technology can utilize it because of its user-friendly design, which leads them through the process. The InceptionV3 model with a greater accuracy result was applied and integrated to the application for classifying the diseases from rice leaf images.

Moreover, the program incorporates a real-time video capture feature, going beyond static image input. Using the camera on their device, users can record live footage of rice leaves and receive instantaneous predictions for dynamic crop health monitoring. For prompt interventions and proactive disease management, this real-time component is crucial. The Tkinter application uses the trained deep learning model to accurately predict the disease class of the rice leaves after processing the input. The forecasts are then converted into practical advice suited to the particular ailment found. These suggestions are a great tool for farmers, helping them to make well-informed choices about crop care and management techniques.

## IV. RESULTS

Thus, the application was able to make correct predictions with the help of the deep learning architecture used in the script that is based on the Inception V3 model with transfer learning, augmented by additional layers including dropout and dense layers for rice disease classification.

### A. MODEL ACCURACY

The model gained an impressive accuracy of approximately 96% on training and 87.50% on validation set with Inception V3 model and similarly, 100% and 87.50% accuracy on mobileNetV2 which was a excellent result to deploy model with the application. The result accuracies obtained is tabulated in table 2.

**Table 2.** Applied Models accuracy comparison

| Model used | Training accuracy | Validation accuracy |
|---|---|---|
| ResNet-50 | 87% | 75% |
| DenseNet121 | 100% | 90% |
| VGG16 | 76% | 84% |
| MobileNetV2 | 94% | 95.83% |
| Inception V3 | 96% | 86% |
| EfficienNetB0 | 33% | 33% |
| ResNet-101 | 58% | 54% |
| VGG19 | 98% | 70% |
| NasNet | 100% | 80% |
| DenseNet169 | 100% | 91% |

Thus, we integrated MobileNetv2 model by analyzing the model loss and training and validation accuracy results which were satisfying enough. Mass overfitting was seen on VGG19, NasNet and low performance on efficientNetB0 as seen in table 2.

As, shown in figure.5. the inception DenseNet121 model obtained excellent result on training and validation set with the approach.

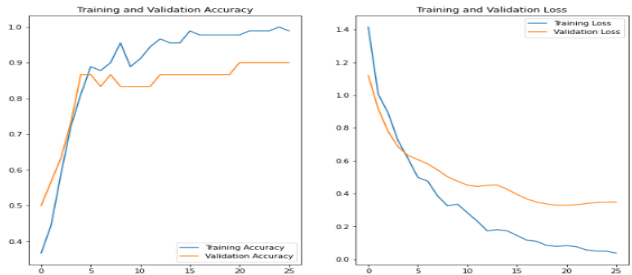

**Figure 5**. History plot for Desenet121

The accuracy of training and validation over time is displayed on the left graph. With a range of 0 to 1, the "Accuracy" y-axis is labeled. "Epoch" is the label for the x-axis, which has a range of 0 to 8. The graph consists of two lines: a green line that reads "Validation Accuracy" and a blue line that reads "Training Accuracy." Over time, both lines rise from their initial values of about 0.5. At the end of the graph, the validation accuracy line reaches a value of approximately 0.85, while the training accuracy line reaches a value of approximately 0.9. At the end of the graph, the validation loss line reaches a value of roughly 20, while the training loss line reaches a value of roughly 0. The training and validation loss over time are displayed in the graph on the right. "Loss" is the label on the y-axis, which runs from 0 to 80. "Epoch" is the label for the x-axis, which has a range of 0 to 8. The graph consists of two lines: a green line that reads "Validation Loss" and a blue line that reads "Training Loss." At first, both lines are about 80, and they gradually get shorter. At the end of the graph, the validation loss line reaches a value of approximately 20, while the training loss line reaches a value of approximately 0.

The 10 different models used on small dataset shows comprehensive exploration on a small dataset for rice leaf obtained from UCI repository. The graphs as a whole provides the insights that the deep learning model is training successfully. Both the training accuracy and the training loss are rising. Not as high as the training accuracy, but still rising, is the validation accuracy. This shows that the model can effectively generalize to new data. As, seen in figure 3. The model gained an accuracy of about 96% on training and resulted in a good prediction on unseen image data. The blue line represents the accuracy of the model on a validation set, and the red line shows the cross entropy (a measure of how well the model is predicting the correct class). The accuracy appears to be increasing as the number of epochs increases, which is a good sign. The cross entropy is also decreasing, which is another good sign. Overall, the image shows that the model is learning and improving over time. The cross entropy of the model is about 0.60.

### B. PREDICTION TEST

The model could successfully predict the type of disease leaf is associated or affected with and GUI application was made in such a way to provide recommendations to the user on the type of disease affected and solutions to overcome them.

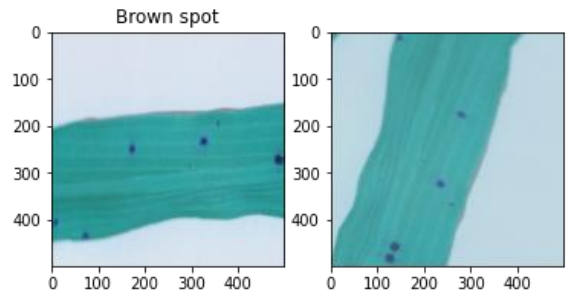

**Figure 6.** Brown spot disease classification

The figure 6 shows prediction of brown sport disease on a sample test image. The sort of disease affecting the rice leaves has been successfully predicted by integrating the trained model into the GUI application. Users can upload an image of a diseased rice leaf using the GUI's user-friendly interface, and the model quickly predicts the associated disease label. The remarkable precision of the forecasts enables farmers and users to promptly recognize possible problems impacting their crops.

The application's goal and theme are reinforced by the classification.jpg image, which acts as a standout logo at the top of the graphical user interface. The resized recycle.png image and the logo are two examples of the visual components that make up a visually appealing and well-designed user interface.

### C. MODELS HYPERPARAMETERS

Deep learning models' training dynamics and performance are greatly influenced by their hyperparameters. During optimization, the number of steps taken to minimize the loss function is controlled by the learning rate. Selecting the right learning rate is essential; too high a rate could push the model toward an unsatisfactory solution too soon, while too low a rate could result in a delayed and possibly halted convergence. The amount of training samples needed to calculate each gradient update depends on the batch size. The various hyperparameters used are shown in table 3.

**Table 3.** Common hyperparameters used for models

| Hyperparameter | Description | Value(s) |
|---|---|---|
| Learning Rate | Controls the step size during optimization | Not explicitly set; uses |

|  |  | Adam's default rate (0.001) |
|---|---|---|
| Batch Size | Number of samples per gradient update | 128 |
| Number of Epochs | Number of complete passes through the training data | 100 |
| Optimizer | Algorithm for updating model weights | Adam |
| Dropout Rate | Fraction of neurons to drop during training | 0.1 |
| Activation Function | Function applied to neurons to introduce non-linearity | ReLU (for hidden layers), Softmax (for output layer) |

While a larger batch size expedites training, it may also raise the danger of overfitting if the model is too dependent on the training set. A smaller batch size may result in noisier updates but may enhance generalization. The number of epochs determines how many times the whole dataset will be used to train the model. The model can learn more efficiently with more epochs, but overtraining can cause overfitting. We utilized the early stopping process for the best outcome.

The algorithm called Optimizer is in charge of adjusting the weights of the model by using the gradients that were calculated during training. Weight changes are handled differently by various optimizers, such as Adam, SGD, and RMSprop, which affects the convergence behavior and effectiveness of the model.

Dropout Rate is a regularization strategy that reduces overfitting by randomly ignoring a portion of neurons during training. This keeps the model from growing overly dependent on any one neuron. Last but not least, the model gains non-linearity from the activation function, which enables it to recognize and interpret intricate patterns in the data. The way the model interprets and learns from inputs is determined by functions like ReLU, Sigmoid, and Tanh, which has a big impact on the model's capacity to grasp complex correlations in the data. The performance of the model and generalization must be balanced by carefully adjusting each hyperparameter.

### D. TKINTER APPLICATION

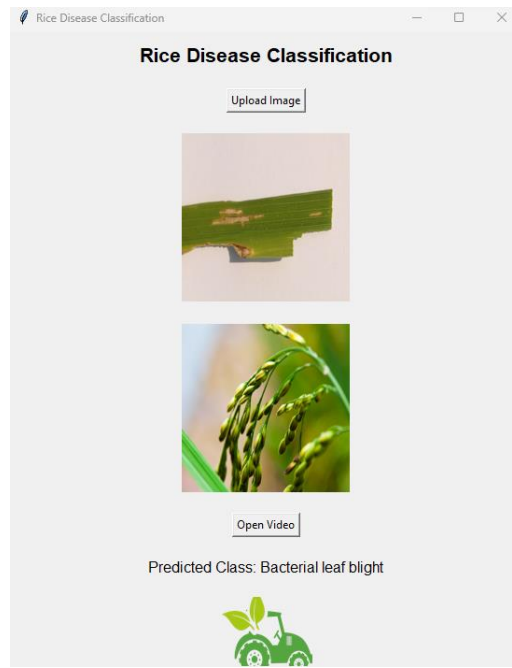

**Figure 6.** Application GUI

Finally, the GUI was able to integrate the ML model, and then correct predictions were made and classification of disease could be done successfully. The recommendations could be provided to the farmers successfully with the help of the application. An important step toward bridging the gap between cutting-edge technology and useful agricultural applications has been taken with the successful integration of the GUI with the machine learning model. Interaction between the user and the system's underlying intelligence is made possible by the interface's seamless integration with the deep learning model. This integration highlights the potential of technology to empower people in traditionally non-technical domains while also simplifying the user experience. Based on the MobileNetV2 model architecture [17], the deep learning model continuously processes the input data as the user works with the GUI, uploading photos or recording videos of rice leaves. By utilizing transfer learning, the model makes use of its prior knowledge from ImageNet to identify complex patterns and features that correspond to different rice diseases. The process's successful completion demonstrates the deep learning architecture's versatility and resilience in the context of classifying agricultural diseases [18], [19]. The transfer learning algorithm shows a great result for promoting rice leaf prediction with intelligent application [27]. The application GUI built is shown in figure 7.

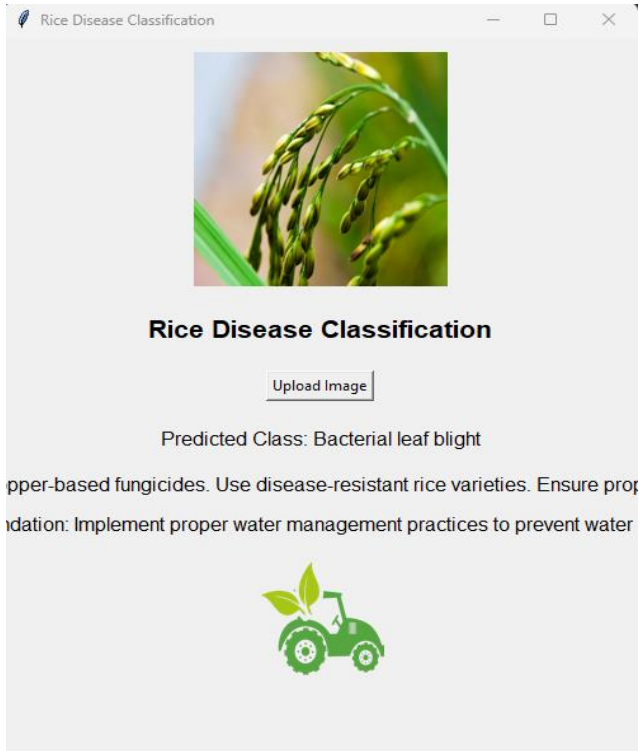

**Figure 7.** GUI application final

## V. DISCUSSION

The accurate diagnoses and detailed categorization of rice illnesses represent the culmination of this integration. The model's capacity to learn from and generalize from a variety of data is demonstrated by its accuracy in identifying and labeling diseases, which empowers it to make judgments about crop health [18]. For farmers looking for trustworthy information to direct their farming practices and lessen the impact of diseases on crop yield, this accuracy is crucial [19]. The result obtained from this research study has been compared with previous research works, as seen in Table 4.

**Table 4.** State of the art comparison with previous works

| Study | Methodology | Validation Accuracy | Outcome |
|---|---|---|---|
| [1] | Ensemble Model | 91% | Effective field diagnosis, reasonably good accuracy |
| [2] | Image Processing with ML | - | Faster, more accurate than traditional methods |
| [3] | Ensemble Classification with Feature Selection | - | Efficient disease prediction model |
| [4] | 10-Fold Cross-Validation with ML | 94% | High accuracy, accessible to developing nation |
| [5] | Deep Learning & ML Comparison | 78.2% (Best DL Model) | DL models outperform ML, 5-layer CNN is best |
| [6] | Deep Learning Ensemble Model | 91% | High accuracy, smartphone app deployment |
| [7] | CNN-Based Model | 95% | Successful classification, robust to different conditions |
| [8] | Transfer Learning with InceptionResNetV2 | 95.67% | High accuracy with transfer learning |
| [9] | Quadratic SVM with Feature Extraction | 81.8% | Satisfactory accuracy with shape criteria |
| [10] | Transfer Learning with CNN-VGG19 | 93% | Superior performance compared to baseline models |
| [11] | Deep Learning & ML Comparison | 78.2% (Best DL Model) | DL outperforms ML, 5-layer CNN is best |

| [12] | YOLOv8 AI-based Two-Stage Method | 89.9% | Higher accuracy than YOLOv7 and YOLOv5 |
|---|---|---|---|
| Our results | Transfer learning algorithms used: VGG16. DenseNet121, Resnet 50, MobileNet v2 & inception V3 | DenseNet121 (90%) and MobileNet V2 (95%) excelled in our study, while ResNet-50 (75%) and VGG16 (84%) underperformed including overfitting on 2 models. | DenseNet 121 (90%) and MobileNetV2 (92%) excelled in our study, while ResNet-50 (75%) and VGG16 (84%) underperformed. |

One of major achievement of this research work is a remarkable accuracies of 90%+ on 3 different transfer learning models on such a small dataset comprising of 120 images. The GUI built with in tkinter goes beyond simply classifying diseases; it can also offer farmers helpful recommendations [26]. These suggestions are based on the model's forecasts and are made specifically to deal with the identified disease. With the help of this feature, the application becomes more than just a classification tool—rather, it becomes a dynamic decision support system that gives farmers useful insights for managing their crops. The fact that the recommendations were delivered successfully demonstrates how well sophisticated machine learning and approachable user interfaces can be combined to solve practical agricultural problems [7]. The work has several limitations like unsuitability for integrating to mobile devices, and application isnt suited for high end Pcomputer devices as well as mobile devices, in future we aim to minimize such issues. The data can be further enhanced in the future with more data collection and utilization for making more robust models and developing mobile application tools. Many works have been done in deep learning in the agriculture sector [15], along with transfer learning models [13], process models to have been used [14], along with federated learning processes [23]. AI and robotics have a great role in promoting sustainability with multidisciplinary approaches like waste management aid [30], [31], and disease classifications [22-29]. There is still a need for much technological advancement in the field of agriculture, and the development of these applications can be supportive to create great outcomes in agriculture productions [28].

## VI. CONCLUSIONS

The GUI's seamless integration with the machine learning model demonstrates the architecture's technological prowess and is a big step toward enabling farmers to use cutting-edge technologies. This development with analysis on 10 models shows how crop illnesses may be accurately diagnosed by deep learning and actionable advice can be given, leading to more productive and sustainable farming methods.

With an AI-integrated system, it can be used to scan rice leaves and predict (classify) the disease. In addition to offering a video classification feature for a range of agricultural applications, the GUI makes it simple for users to upload photos for classification. Farmers are empowered to make informed decisions about crop management due to accurate disease categorization, which is made possible by the smooth interface between the model and GUI.

## AUTHOR BIOGRAPHY

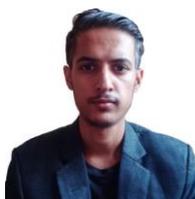

**Biplov Paneru** is currently pursuing a Bachelor of Engineering degree in Electronics and Communications at Nepal Engineering College, which is affiliated with Pokhara University. His academic journey has been supplemented by a strong involvement in R&D initiatives, particularly in computer vision, embedded systems, and image processing. Along with his studies, Biplov works as a dedicated rocket research and development engineer at Nepal's National Innovation Center, where he contributes his expertise to propelling advancements in aerospace technology. Furthermore, he works as a freelance software developer on Upwork, demonstrating his versatility in software development projects across multiple domains. For being familiar to his more scholarly works refer to: https://orcid.org/0000-0003-2003-0648

**Bishwash Paneru**

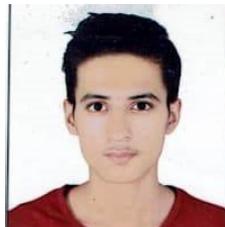

is currently pursuing a Bachelor of Engineering degree in Chemical Engineering at the Institute of Engineering, Puclhowk Campus. Throughout his academic career, he has shown a strong interest in R&D projects, focusing on green hydrogen, pyrolysis, and water treatment. As a research intern at Environ Renewables, Bishwash actively participates in projects focused on green hydrogen and alternative energy sources, demonstrating his commitment to sustainability and renewable technologies. He uses his experience as a chemical engineering freelancer on Upwork to work on a variety of projects related to his field of study. Bishwash's dedication to pushing the boundaries of chemical engineering through research and practical applications demonstrates his desire to contribute to innovative solutions for energy and environmental sustainability. Orcid: https://orcid.org/0009-0000-4025-2542

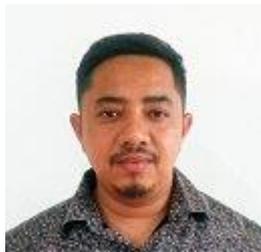

**Krishna Bikram Shah** is currently an assistant professor in the Department of Computer Science and Engineering at Nepal Engineering College, which is affiliated with Pokhara University. He also serves as a visiting faculty member at Academia International College in Lalitpur, which is affiliated with Tribhuvan University. Mr. Shah's research interests are broad, spanning fields such as Artificial Intelligence (AI), Remote Sensing and Geographic Information Systems (RS-GIS), Digital Image Processing (DIP), Bioinformatics, Artificial Intelligence of Things (AIoT), and Solid Waste Management. His contributions cover a wide range of academic endeavors, making him a well-known figure in his field. For a thorough understanding of his scholarly work, please see his ORCID profile at https://orcid.org/0000-0003-1763-511X.